\documentclass[10pt]{article} 
\usepackage[accepted]{tmlr}

\usepackage[utf8]{inputenc} 
\usepackage[T1]{fontenc}    
\usepackage{hyperref}       
\usepackage{url}            
\usepackage{booktabs}       
\usepackage{amsfonts}       
\usepackage{nicefrac}       
\usepackage{microtype}      
\usepackage{xcolor}         
\usepackage{graphicx}
\usepackage{amsmath}
\usepackage{multirow}
\usepackage{enumitem}
\usepackage{siunitx}
\usepackage{caption}
\usepackage[table]{xcolor}
\usepackage{colortbl}
\usepackage{wrapfig}
\usepackage{listings}
\usepackage{tabularx}
\usepackage{listings}
\definecolor{addblue}{RGB}{0,90,180}

\title{Formatting Instructions for TMLR \\Journal Submissions}

\definecolor{highlightblue}{HTML}{E8F0FE}
\definecolor{highlighttan}{HTML}{FDF2D0}
\definecolor{addblue}{HTML}{1A56A0}
\title{SATFormer: Learning When to Use Early Value Residuals in Transformers}
\author{%
  \name Skye Gunasekaran
  \email akgunase@ucsc.edu \\
  \addr UC Santa Cruz
  \AND
  \name Téa Wright
  \email teaywright@berkeley.edu \\
  \addr UC Berkeley
  \AND
  \name Rui-Jie Zhu
  \email rzhu48@ucsc.edu \\
  \addr UC Santa Cruz
  \AND
  \name Jason Eshraghian
  \email jeshragh@ucsc.edu \\
  \addr UC Santa Cruz
}
\def\openreview{\url{https://openreview.net/forum?id=flq3oO3cNr}}
\begin{document}

\maketitle

\begin{abstract}

Recent Transformer architectures expose later layers to representations computed in earlier layers, motivated by the observation that low-level features can become harder to recover as the residual stream is transformed through depth. One such method adds static value residuals: learned mixing coefficients that expose the first-layer value projection $V_1$ uniformly across tokens and heads. However, we theorize that the usefulness of $V_1$ is unlikely to be constant across tokens, heads, and contexts, and therefore treat early-representation reuse as a retrieval problem. Here, we introduce \textbf{Selective Access Transformer (SATFormer)}, which preserves the first-layer value pathway while controlling access with a context-dependent gate. Across models from 130M to 1.3B parameters, SATFormer improves validation loss over the static value-residual and Transformer baselines at every evaluated scale, and improves average zero-shot accuracy over both at all four scales. Its strongest gains appear on retrieval-intensive benchmarks, where it improves over static value residuals by approximately 1.5 average points, while maintaining near baseline throughput and memory. Mechanistic analyses suggest sparse, depth-dependent, head-specific, and category-sensitive access patterns, supporting the interpretation that SATFormer learns selective reuse of early representations rather than uniform residual copying. Our code is available at \href{https://github.com/skyegunasekaran/satformer}{https://github.com/skyegunasekaran/satformer}.
\end{abstract}

\section{Introduction} 
Information flow across depth has motivated a broad family of architectural modifications to the Transformer. Although the residual stream provides a direct pathway for propagating token representations, each layer transforms those representations through attention, normalization, and feed-forward computation. Prior work has argued that low-level lexical, structural, and token-specific features computed in early layers can become difficult to recover deeper in the network \citep{zhou2025value}. A natural response is to expose later layers to earlier representations, specifically the first-layer value projection $V_1$, through additional cross-layer pathways. This idea underlies work on residual scaling and normalization, gated residuals, dense layer aggregation, and dynamic cross-layer routing \citep{bachlechner2021rezero,touvron2021going,wang2024deepnet,srivastava2015highway,pagliardini2024denseformer,xiao2025muddformer,zhu2024hyper}.

These methods largely share an underlying framing: cross-layer reuse is a connectivity problem, where the design question is how many pathways to add between layers and how richly to combine them. The simplest instantiation, value residual learning (ResFormer) \citep{zhou2025value}, exposes $V_1$ to all subsequent layers through a learned scalar coefficient per layer. This mechanism is computationally cheap, but it applies the same amount of $V_1$ to every token and every attention head within a layer. More expressive methods such as DenseFormer, MUDDFormer, and HyperConnections recover finer-grained access by aggregating over wider layer histories or generating dynamic dense pathways, but these gains come with higher memory cost and lower throughput \citep{pagliardini2024denseformer,xiao2025muddformer,zhu2024hyper}.

This leaves a more targeted question: can a single early-value pathway become competitive if the model learns \emph{control} over when and where that pathway is used? The usefulness of $V_1$ should vary across tokens, heads, layers, and contexts: a token that depends on prompt-specific lexical or semantic information may benefit from direct access to early features, while another token or head may not need that access at all. A layer-wise scalar cannot express this variation, because it exposes or suppresses $V_1$ broadly across the layer. Following ResFormer, we use the first-layer value representation $V_1$ as the fixed source, allowing us to isolate the effect of replacing static layer-wise mixing with input-dependent token- and head-level control.

To put this into practice, we introduce \textbf{Selective Access Transformer} (\textbf{SATFormer}), in which a computationally cheap per-token, per-head gate modulates how much of $V_1$ enters each layer's value computation. Empirically, this control-based design yields favorable performance-efficiency trade-offs. Across 130M-1.3B parameter models, SATFormer improves validation loss over the Transformer and ResFormer baselines at every evaluated scale, and with a thin-and-long variant, improves average zero-shot accuracy over both at all four scales. The gains are particularly visible on retrieval-intensive evaluations, where SATFormer outperforms ResFormer by nearly 1.5 average points. At the same time, SATFormer maintains throughput and memory usage close to the Transformer and ResFormer baselines, while improving on their perplexity at all scales. We further analyze the learned gating behavior to understand how SATFormer uses its retrieval pathway.

Our contributions are as follows:

\begin{itemize}
    \item \textbf{A control-based framing for early-representation reuse.} We treat $V_1$ access as a question of control across tokens, heads, and depth over a single early pathway, rather than connectivity over many pathways.

    \item \textbf{The SATFormer architecture.} SATFormer replaces ResFormer's per-layer scalar with a per-token, per-head gate from a single linear projection of the current hidden state, adding negligible parameters and runtime cost.

    \item \textbf{Pareto-favorable scaling and retrieval.} SATFormer improves validation loss over Transformer and ResFormer at every evaluated scale (130M--1.3B), with the largest zero-shot gain over ResFormer at 1.3B ($+0.67$ average accuracy). On retrieval-intensive benchmarks, it gains approximately 1.5 points over ResFormer and matches the best dense alternative (MUDDFormer) to within $0.01$ average points at roughly $1.8\times$ its throughput.

    \item \textbf{Mechanistic evidence for structured access.} Gate analyses reveal sparse, depth-dependent, head-specific, and token-specific access, concentrated in a subset of heads. Ablation experiments show that both the $V_1$ pathway and the input-dependent structure of the gate are functionally load-bearing for language-modeling performance, and a logit-lens probe is consistent with this picture.
\end{itemize}

\section{Related Work}
\label{2 Related Work}

Prior work improves depth-wise information flow by changing either the stability of the residual stream or the amount of cross-layer history available to later computation. Residual scaling and normalization methods such as ReZero \citep{bachlechner2021rezero}, LayerScale \citep{touvron2021going}, and DeepNorm \citep{wang2024deepnet} improve optimization and signal propagation without adding explicit access to a large layer history. Gated residual mechanisms such as Highway networks \citep{srivastava2015highway} similarly introduce control over residual updates. A second line of work increases cross-layer access more directly: multi-stream recurrence methods such as HyperConnections \citep{zhu2024hyper} and mHC \citep{xie2025mhc} expand the residual stream into multiple pathways, while aggregation and routing methods such as DenseNet-style connectivity \citep{huang2017densely}, ELMo-style learned layer mixtures \citep{peters-etal-2018-deep}, DenseFormer \citep{pagliardini2024denseformer}, MUDDFormer \citep{xiao2025muddformer}, MRLA \citep{fang2023cross}, LAuReL \citep{menghani2024laurel}, and Attention Residuals \citep{team2026attention} combine information from earlier layers more explicitly. These methods trace an efficiency--granularity frontier: static scalar connections are inexpensive but coarse, whereas dense aggregation and dynamic routing provide finer-grained access at higher memory cost and lower throughput. SATFormer targets a different point on this frontier: adaptive control over a single early-value pathway, rather than denser access to many previous representations.

\begin{figure}
    \centering
    \includegraphics[width=0.95\linewidth]{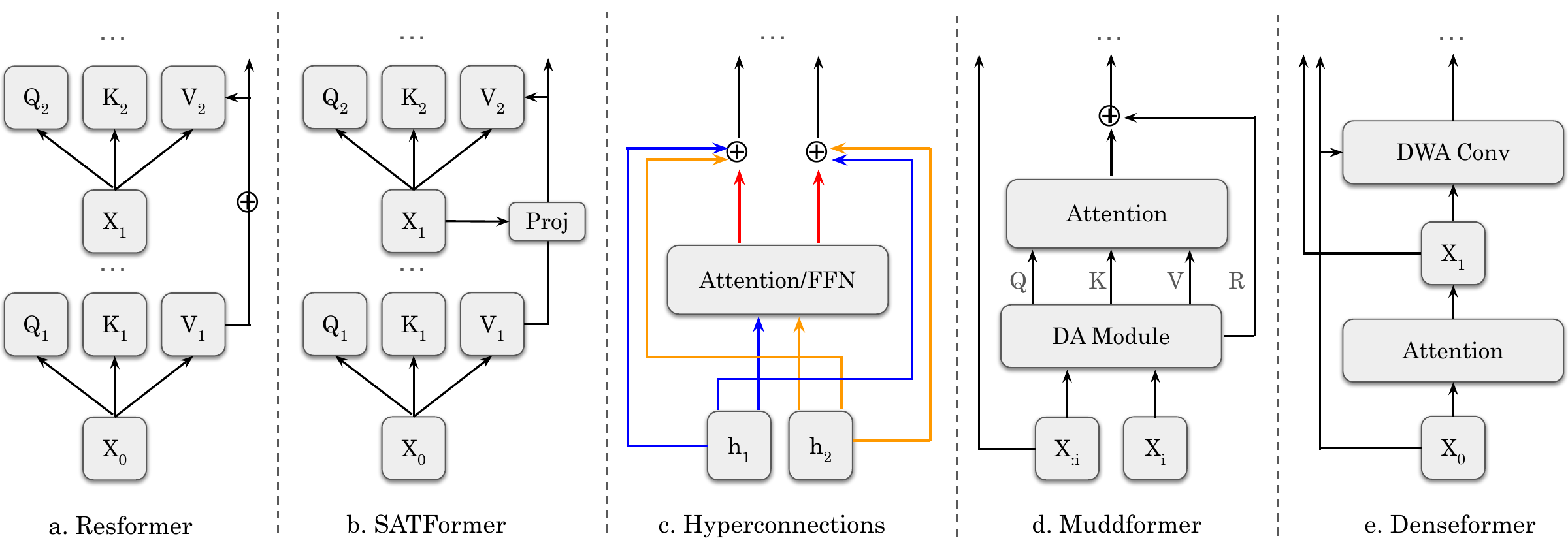}
    \caption{\textbf{SATFormer increases access granularity without increasing connection density.} ResFormer provides efficient but static layer-wise access to $V_1$; DenseFormer, MUDDFormer, and HyperConnections use denser or more dynamic pathways at substantially higher cost. SATFormer instead preserves ResFormer's single early-value pathway and makes access token- and head-dependent through a lightweight gate.}
    \label{figure1}
\end{figure}

The closest prior work to SATFormer is value residual learning (ResFormer) introduced by \citep{zhou2025value}. ResFormer addresses the dilution of initial token information by adding a direct connection from the first-layer value projection to the value vectors of subsequent layers. For a Transformer layer $n > 1$, the modified value vector is computed as

\begin{equation}
\label{equation1}
V^{\prime(n)} = \lambda^{(n)}_{1} V^{(1)} + \lambda^{(n)}_{2} V^{(n)}
\end{equation}

where $V^{(1)}$ is the value representation from the first layer, $\lambda^{(n)}_{1}$ is typically a learned layer-wise scalar initialized to $\lambda_{\text{scale}} \cdot \frac{e^{\lambda'}_{n,1}}{\sum^{N}_{j=1} e^{\lambda'}_{j,1}}$ where $\lambda_{\text{scale}}$ is initialized to the total layers $N$ and $\lambda^{(n)}_{2}$ is held constant across layers. Note, in the definition above we refer to the highest performing ResFormer variant, Learnable ResFormer Plus. 

We focus our empirical comparisons on four methods that instantiate distinct points along the efficiency–granularity frontier. Value residual learning \citep{zhou2025value} addresses the dilution of initial token information by adding a direct residual connection to the value vectors of all subsequent layers, using learned layer-wise scalar coefficients. HyperConnections \citep{zhu2024hyper} provides a learnable alternative to traditional residuals by replicating hidden states into multiple pathways. MUDDFormer \citep{xiao2025muddformer} extends this to the most flexible setting, generating dense connection weights dynamically for each sequence position via a lightweight MLP. DenseFormer \citep{pagliardini2024denseformer} introduces a Depth-Weighted-Average module after each Transformer block, computing a weighted average over all preceding representations.

Attention Residuals \citep{team2026attention} restructures the residual stream into an attention-weighted combination across layers and reports that per-head granularity hurts in that setting. This does not directly contradict SATFormer: Attention Residuals partitions the weighting of an entire cross-layer residual mixture, whereas SATFormer keeps the standard residual stream and uses KV-head-specific control only for a single, fixed value source. Gated Attention \citep{qiu2026gated} is closer in form but acts at a different location, applying a head-specific sigmoid gate to the SDPA output before the output projection. Appendix~\ref{appendix:bgate} tests whether this output gate already subsumes selective value access.

SATFormer targets a narrower and more controlled design point. Rather than aggregating over all previous layers, expanding the residual stream, or routing across a dense layer history, it asks whether a single early-value pathway can become competitive when access to that pathway is made input-dependent. SATFormer does not increase the amount of cross-layer history available to the model; it increases the model's control over when and where one early representation is reused.

\paragraph{Why control the value stream?}
Value-stream gating provides a lightweight and interpretable intervention on the same \(V_1\) pathway used by ResFormer: a single \(d_{\mathrm{model}}\!\rightarrow\!N_{kv}\) projection adds negligible overhead while isolating changes in the access policy. The learned gates are sparse and depth-structured, and ablating either the pathway or its input dependence worsens perplexity (Figures~\ref{figure3},~\ref{figure4},~\ref{figure8}), indicating that the mechanism is selectively used and functionally important. Appendix~\ref{appendix:bgate} further shows that selective value access improves perplexity even when an attention-output gate is present, suggesting that the two mechanisms are complementary rather than redundant.
\section{Methodology}
\label{3 Methodology}

Previous work exposes $V_1$ through layer-wise scalar coefficients, making early-value access uniform across tokens and heads within a layer \citep{zhou2025value}. To enable more dynamic input-aware gating, we replace the static scalar $\lambda^{(n)}_{1}$ in Equation~\ref{equation1} with an input-dependent mixing coefficient. This allows the model to perform \textit{context-aware value retrieval}. Let $x_t^{(n)} \in \mathbb{R}^{d_{\mathrm{model}}}$ denote the normalized hidden state for the token at position $t$ in layer $n$. SATFormer computes a per-token, per-KV-head gate $\alpha_{t,j}^{(n)}$ as

\begin{equation}
\label{equation2}
\alpha_{t,j}^{(n)} =
\left[\operatorname{ReLU}\!\left(x_t^{(n)} W_{\alpha}^{(n)}\right)\right]_j,
\end{equation}

where $W_{\alpha}^{(n)} \in \mathbb{R}^{d_{\mathrm{model}} \times N_{kv}}$ is a single linear projection, $N_{kv}$ is the number of key-value heads, and $j \in \{1,\ldots,N_{kv}\}$ indexes the KV head. The value used by attention is then

\begin{equation}
\label{equation3}
V_{t,j,r}^{\prime(n)} = V_{t,j,r}^{(n)} + \alpha_{t,j}^{(n)} \cdot V_{t,j,r}^{(1)},
\end{equation}

where $r$ indexes the per-head value dimension. The $\operatorname{ReLU}$ activation keeps the mixing coefficient non-negative and allows the gate to become exactly zero. A deeper comparison of gating mechanisms can be found in Appendix~\ref{appendix:b1}.

\paragraph{Design rationale.} SATFormer makes three deliberately minimal choices. \textbf{First}, the gate is per-token and per-head because the usefulness of $V_1$ varies along both axes: a token-level gate alone would ignore head-level specialization, while a head-level gate alone would make access uniform across the sequence, preventing the model from selecting particular tokens whose early representations should be preserved. \textbf{Second}, the gate is computed with a single linear projection from the current normalized hidden state. This makes access depend on the representation already available to the layer, while adding only $d_{\mathrm{model}}N_{\mathrm{kv}}$ parameters per layer. \textbf{Third}, we use a ReLU gate so that access is non-negative and can be exactly zero. Non-negativity keeps the pathway interpretable as additive reuse of the first-layer value stream, while exact zeros allow the model to fully disengage from $V_1$ when early information is unhelpful. Further ablation studies regarding gating functions can be found in Appendix \ref{appendix:b1}. This construction preserves the same early-value source as ResFormer, but changes the access policy: rather than applying a single layer-wise coefficient, SATFormer lets each token and KV head choose how much of $V_1$ to reuse.

\subsection{ReLU Gating}

\textbf{Gate sparsity.} Equation~\ref{equation2} permits exact zeros, so the architecture can close the $V^{(1)}$ pathway for an individual token--head pair. This is an architectural possibility rather than an assumed outcome: empirically, Figure~\ref{figure3} and Figure~\ref{figure8} show many zero or near-zero gates in early layers, followed by stronger and more head-specialized access later in depth.

\textbf{Direct cross-layer gradient flow.} Let $G_{t,j}^{(n)}=\partial\mathcal{L}/\partial V_{t,j}^{\prime(n)}$. Holding the gate fixed to isolate the direct shortcut term, the contribution of later layers to the gradient of the first-layer value is
\begin{equation}
\label{equation4}
\left.\frac{\partial \mathcal{L}}{\partial V_{t,j}^{(1)}}\right|_{\mathrm{shortcut}}
= \sum_{n>1} \alpha_{t,j}^{(n)} G_{t,j}^{(n)}.
\end{equation}
If $V^{(1)}=X^{(1)}W_V^{(1)}$, this term is propagated to $W_V^{(1)}$ by the usual linear-layer Jacobian. Thus, a zero gate removes the direct shortcut contribution for that token--head--layer triple, while a positive gate scales it. SATFormer therefore learns the strength and sparsity of its direct cross-layer gradient routes instead of fixing them uniformly.

\textbf{Unbounded gates do not create recurrent amplification in this design.} ReLU is unbounded above, so boundedness cannot be guaranteed by construction. Gate inputs are normalized hidden states, and all experiments use gradient clipping at 1.0. Empirically, the gate statistics remain depth-structured rather than escalating (Figure~\ref{figure8}), and we observed no gate-related loss spikes or divergence from 130M to 1.3B parameters. The bounded sigmoid and tanh alternatives also converge to worse loss in Appendix~\ref{appendix:b1}.

\section{Experiments}
\label{4 Experiments}

We organize our evaluation around the central claim tested in this work: under our pretraining setup and against the evaluated baselines, SATFormer provides a favorable efficiency-performance trade-off with gains on the retrieval benchmarks considered here and on language-modeling loss. We therefore begin with retrieval-intensive benchmarks, where Selective Access Transformer should be most directly useful. We then evaluate wall-clock efficiency, throughput, and memory usage to test whether SATFormer preserves the computational profile of sparse connectivity. Finally, we report standard language modeling, zero-shot, and scaling results to verify that these targeted gains do not come at the cost of model quality.

\paragraph{Pretraining setup.}
Along with SATFormer, our experiments include Transformer, ResFormer, DenseFormer, MUDDFormer, and HyperConnections as baselines. We train all models under matched data, optimizer, and hardware conditions across three main scales: Small (130M parameters, 5B tokens, $\eta=10^{-3}$), Medium (340M parameters, 10B tokens, $\eta=1.5 \times 10^{-3}$), and Large (760M parameters, 20B tokens, $\eta=2 \times 10^{-3}$). We additionally train Transformer, ResFormer, and SATFormer at the XL scale (1.3B parameters, 30B tokens) to study scaling relative to the closest static value-residual baseline. All models are trained on 2 NVIDIA H200 GPUs, with configurations closely following \citet{yang2024gated}.

We train on a randomly sampled subset of FineWeb-Edu and use the Llama-2 tokenizer with a vocabulary size of 32,000. Training uses a sequence length of 4K tokens and the AdamW optimizer with weight decay 0.1 and gradient clipping 1.0. The learning rate $\eta$ follows a cosine annealing schedule, decaying to a minimum of $0.1 \times \eta_{base}$ after a linear warmup period covering the first 1\% of training tokens.

\paragraph{Model configurations.}
\begin{wraptable}{r}{0.45\textwidth}
    \centering
    \small
    \scshape
    \vspace{-15pt}
    \caption{\textbf{Model Configurations.}}
    \label{table1}
    \begin{tabular}{lcccc}
        \toprule
        \textbf{Variant} & \textbf{$d_{model}$} & \textbf{$L$} & \textbf{$n_{heads}$} & \textbf{$d_{ff}$} \\
        \midrule
        Small  & 768  & 11 & 12 & 3072 \\
        Medium & 1024 & 18 & 16 & 4096 \\
        Large  & 1536 & 19 & 24 & 6144 \\
        XL     & 2048 & 24 & 32 & 5632 \\
        \bottomrule
    \end{tabular}
    \vspace{-5pt}
\end{wraptable}
Table~\ref{table1} reports the model configurations used at each scale. For each baseline, we use the strongest or recommended variant from the corresponding prior work. DenseFormer uses a dilation factor of 4 and a Depth-Weighted-Average (DWA) period of 5. MUDDFormer uses full Query, Key, Value, and Residual connectivity with parameter reallocation and pre/post Depth Aggregation normalization. HyperConnections uses the dynamic connection variant with expansion rate $n=2$, tanh activation, and layer normalization. ResFormer follows the Learnable ResFormer Plus variant defined by the weighted residual sum in Equation~1.

\subsection{Retrieval-Intensive Evaluation}

We first evaluate SATFormer on six retrieval-oriented benchmarks following \citet{arora2024just}, with inputs truncated to 2K tokens, to test whether selective early-value access helps retain prompt-specific information in these settings.  As shown in Table 2, SATFormer attains the best retrieval average among the evaluated architectures, effectively tied with MUDDFormer (29.86 versus 29.86, a $0.003$-point difference we do not treat as a meaningful separation) and improving over ResFormer by approximately 1.5 average points.

This comparison is particularly informative because SATFormer and ResFormer both expose the first-layer value stream. Their main difference is that ResFormer mixes this stream using layer-wise static coefficients, whereas SATFormer conditions access on the current hidden state at the token and head level. The observed improvement therefore suggests that the benefit is not simply due to making $V_1$ available, but may also depend on controlling when and where the early representation is reused.

\begin{table}[htb]
\centering
\caption{\textbf{SATFormer matches the best retrieval average while retaining sparse connectivity.} Its average is effectively tied with MUDDFormer and about 1.5 points above ResFormer. Input length is truncated to 2K tokens; the evaluation protocol follows \citet{arora2024just}.}
\label{table2}
\begin{small}
\begin{sc}
\begin{tabular}{lccccccc}
\toprule
\textbf{Model} & \textbf{TriviaQA} & \textbf{SWDE} & \textbf{SQUAD} & \textbf{NQ} & \textbf{FDA} & \textbf{Drop} & \textbf{Avg}\\
\midrule
Transformer      & 52.9	& 29.43	& 34.91	& 12.06	& 19.14	& 20.07	& 28.085 \\
DenseFormer      & 52.66 & 32.31 & 35.32 & 11.24 & 16.42 & 22.66 & 28.435 \\
MUDDFormer       & 54.32 & \textbf{36.81} & 34.78 & 12.06 & 20.32 & 20.84 & 29.855 \\
HyperConnections & \textbf{55.27} & 25.83	& 36.72	& \textbf{12.82}	& 15.78	& \textbf{24.86} & 28.546 \\
ResFormer        & 53.9	& 27.9 & 34.88 & 11.46 & 19.41 & 22.8 & 28.392 \\
SATFormer   & 54.26 & 32.4	& \textbf{36.96}	& 12.16	& \textbf{21.96}	& 21.41	& \textbf{29.858} \\
\bottomrule
\end{tabular}
\end{sc}
\end{small}
\end{table}
Importantly, SATFormer achieves this retrieval performance while preserving a computational profile close to the baseline Transformer and ResFormer. In contrast, MUDDFormer attains a similar retrieval average but trains substantially more slowly under matched hardware. Within this six-dataset, 2K-token evaluation, these results are consistent with the hypothesis that selective early-value access can help retain prompt-specific lexical or semantic information, while avoiding the higher training cost of the dense baselines evaluated here. One scope caveat applies to this comparison. Every dataset in the protocol of \citet{arora2024just} fits within 2K tokens, so Table~\ref{table2} measures selective access only under short contexts.

\subsection{Needle-in-a-Haystack Evaluations}
\label{sec:niah}
\begin{table}[htb]
\centering
\caption{\textbf{Needle-in-a-Haystack Evaluations.} Accuracy (\%) averaged over interior needle depths (25\%, 50\%, 75\%) under identical YaRN length extension. The $\Delta$ column shows change from 4K to 32K.}
\label{tableA2}
\small
\begin{tabular}{lrrrrrr}
\toprule
\textbf{Model} & \textbf{4K} & \textbf{8K} & \textbf{16K} & \textbf{32K} & \textbf{Combined} & \textbf{$\Delta$ 4K$\rightarrow$32K} \\
\midrule
Transformer       & 68.0 & 61.3 & 36.7 &  7.3 & 43.3 & $-60.7$ \\
DenseFormer       & 71.3 & 70.7 & 71.3 & 59.3 & 68.2 & $-12.0$ \\
MUDDFormer        & \textbf{88.7} & \textbf{91.3} & \textbf{83.3} & \textbf{65.3} & \textbf{82.2} & $-23.3$ \\
HyperConnections  & 50.7 & 51.3 & 12.7 &  0.0 & 28.7 & $-50.7$ \\
ResFormer         & 13.3 &  0.0 & 33.3 &  5.3 & 13.0 & $-8.0$ \\
SATFormer         & 33.3 & 68.0 & 72.0 & 46.0 & 54.8 & $\mathbf{+12.7}$ \\
\bottomrule
\end{tabular}
\end{table}
We separately evaluate single-key retrieval under increasing context length using Needle-in-a-Haystack. Because the models are trained at 4K context and extended beyond 4K using YaRN, we treat this experiment as a targeted stress test of retrieval under context-length extrapolation rather than as evidence of general long-context capability. We avoid composite suites such as RULER \citep{hsieh2024ruler}, which are designed to stress-test frontier-scale models. 

Models are trained at 4K sequence length. Evaluation beyond 4K applies YaRN length extension identically to all architectures. We report retrieval accuracy averaged over the three interior needle depths (25\%, 50\%, and 75\%); depths of 0\% and 100\% are excluded because they are respectively failure- and recency-dominated.

SATFormer overtakes the baseline Transformer at 8K and the gap widens to 46.0 vs.\ 7.3 at 32K. On the combined average across lengths, SATFormer (54.8) exceeds the Transformer (43.3), ResFormer (13.0), and HyperConnections (28.7). SATFormer trails at 4K, and MUDDFormer achieves higher absolute accuracy at every length (DenseFormer does so at all lengths except 16K, where SATFormer is marginally higher). Among the architectures evaluated, SATFormer is the only model with a positive 4K-to-32K endpoint change.
\subsection{Training Efficiency Benchmarks}

\begin{figure}[htb]
    \vspace{-5pt}
    \centering
    \includegraphics[width=0.95\linewidth]{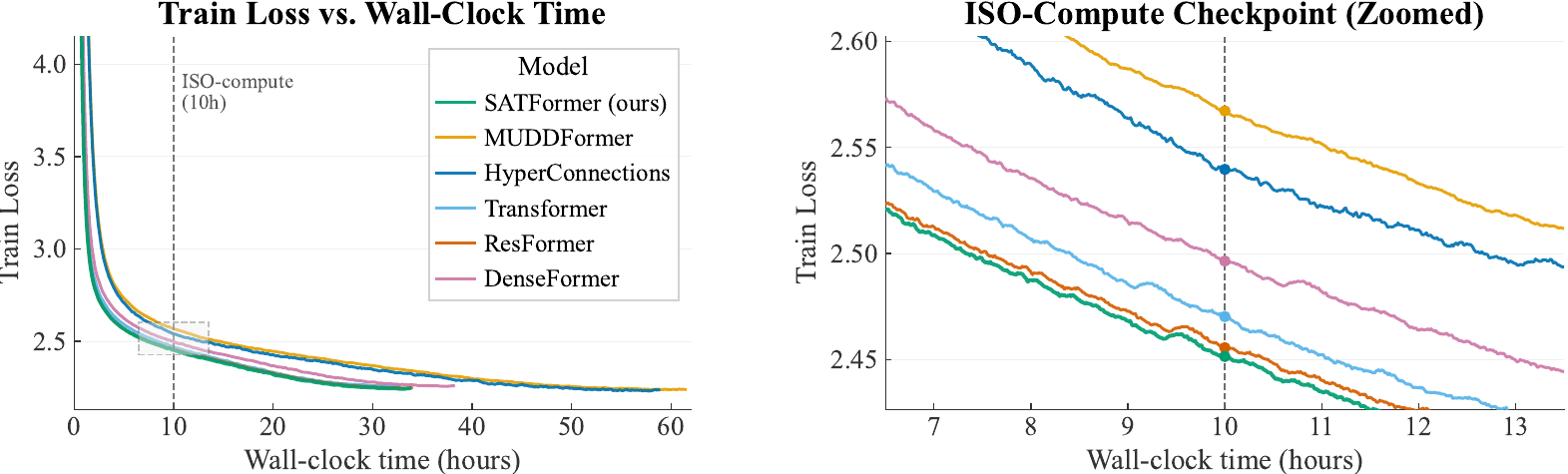}
    \caption{\textbf{At ISO-compute, SATFormer and ResFormer outpace dense cross-layer alternatives by ${\approx}3\times$ wall-clock time.} \textbf{(Left)} Train loss vs.\ wall-clock time; dashed line marks the 10-hour ISO-compute checkpoint. \textbf{(Right)} Zoomed view at 10 hours. All curves are from the 760M training run; SATFormer's advantage over ResFormer is most visible in loss, not wall-clock time, at this scale (2.242 vs.\ 2.246 final validation loss).}
    \label{figure2}
\end{figure}

Final validation loss and zero-shot accuracy do not fully capture the practical cost of dense cross-layer connectivity. For instance, architectures that expand the residual stream, aggregate over many previous layers, or generate dynamic dense pathways may improve endpoint performance while making substantially slower wall-clock progress. We therefore evaluate training efficiency directly, focusing on wall-clock loss, throughput, and memory usage under matched hardware.
\begin{wraptable}{r}{0.45\textwidth}
\vspace{-10pt}
\centering
\caption{\textbf{SATFormer and ResFormer have nearly identical measured training throughput, while dense alternatives incur substantially higher cost.} Tokens/s and peak memory (GB) at the 760M scale.}
\label{table3}
\begin{small}
\begin{sc}
\begin{tabular}{lcc}
\toprule
\textbf{Model} & \textbf{Memory} & \textbf{Throughput} \\
\midrule
Transformer      & 85.76  & 170,308	 \\
DenseFormer      & 93.32  & 145,639 \\
MUDDFormer       & 90.52  & 90,443\\
HyperConnections & 124.73 & 94,534 \\
ResFormer        & 89.59  & 163,826  \\
SATFormer        & 86.03  & 164,839 \\
\bottomrule
\end{tabular}
\end{sc}
\end{small}
\vspace{-10pt}
\end{wraptable}
Figure~\ref{figure2} compares training loss as a function of wall-clock time. At the 10-hour checkpoint, SATFormer achieves a loss of 2.4515 compared to 2.4557 for ResFormer, 2.4703 for the baseline Transformer, and 2.4965 for DenseFormer. In contrast, HyperConnections and MUDDFormer make substantially slower wall-clock progress in this setup, reaching losses of 2.5396 and 2.5672 at the same checkpoint.

Table~\ref{table3} reports memory usage and throughput at the 760M scale with a batch size 16, 4 gradient accumulation steps, sequence length 4096, 2 NVIDIA H200 GPUs, and 5 warmup steps. SATFormer maintains throughput close to the baseline Transformer and ResFormer while using comparable memory. Against the dense alternatives, SATFormer achieves roughly $1.75\times$ higher throughput and uses 31\% less memory than HyperConnections, and $1.82\times$ higher throughput and 5\% less memory than MUDDFormer. Thus, SATFormer provides a more favorable measured wall-clock trade-off than the dense dynamic methods in this setup.

\subsection{Inference Efficiency Benchmarks}
\label{sec:inference}

\textbf{After training}, we re-measure prefill throughput (thousands of tokens per second), decode throughput (tokens per second), and peak GPU memory (GB) at batch size~1 across four sequence lengths. All figures are medians over ten iterations following five warmup iterations, on 2$\times$H200 GPUs with a shared protocol.

\begin{table}[htb]
\centering
\caption{\textbf{Inference efficiency at the 760M scale, batch size 1.} Prefill throughput in ktok/s; decode throughput in tok/s; peak memory in GB. SATFormer closely matches the cheapest baselines (within 6\% of the Transformer on prefill and within 2\% of ResFormer) and remains substantially more efficient than dense cross-layer methods across all sequence lengths.}
\label{tableA1}
\small
\setlength{\tabcolsep}{4pt}
\begin{tabular}{lrrrrrrrrrrrr}
\toprule
& \multicolumn{4}{c}{\textbf{Prefill (ktok/s)}} & \multicolumn{4}{c}{\textbf{Decode (tok/s)}} & \multicolumn{4}{c}{\textbf{Peak mem.\ (GB)}} \\
\cmidrule(lr){2-5}\cmidrule(lr){6-9}\cmidrule(lr){10-13}
\textbf{Model} & 4K & 8K & 16K & 32K & 4K & 8K & 16K & 32K & 4K & 8K & 16K & 32K \\
\midrule
Transformer      & 212.0 & 215.3 & 163.5 & 108.4 & 52.3 & 51.9 & 52.0 & 51.8 & 2.33 & 3.09 & 4.62 & 7.67 \\
DenseFormer      & 187.4 & 194.2 & 152.3 & 101.3 & 46.0 & 45.9 & 45.9 & 46.0 & 2.58 & 3.60 & 5.63 & 9.68 \\
MUDDFormer       &  71.6 &  72.5 &  68.1 &  56.6 & 23.2 & 23.3 & 23.2 & 23.4 & 2.98 & 4.38 & 7.17 & 12.75 \\
HyperConnections & 112.6 & 108.8 &  95.7 &  69.9 & 37.7 & 37.3 & 37.5 & 37.7 & 2.40 & 3.22 & 4.88 & 8.19 \\
ResFormer        & 200.3 & 210.9 & 160.9 & 105.2 & 48.6 & 49.1 & 48.4 & 48.6 & 2.33 & 3.09 & 4.62 & 7.67 \\
SATFormer        & 199.6 & 208.9 & 161.0 & 105.7 & 48.0 & 48.2 & 47.9 & 47.9 & 2.33 & 3.09 & 4.62 & 7.67 \\
\bottomrule
\end{tabular}
\end{table}

\textbf{Memory.} SATFormer's peak memory is indistinguishable from the Transformer and ResFormer at every sequence length ($<$0.1\% difference). The advantage over dense methods widens with context: 39.8\% less memory than MUDDFormer and 20.8\% less than DenseFormer at 32K.

\textbf{Prefill throughput.} SATFormer is within 6\% of the Transformer at 4K; the gap narrows to 2.5\% at 32K. It is essentially identical to ResFormer at all lengths (within 2\%). Against dense alternatives it retains a $1.87\times$ (MUDDFormer) and $1.51\times$ (HyperConnections) advantage at 32K.

\textbf{Decode throughput.} Decode is stable across the $8\times$ range of context lengths for every architecture. SATFormer sustains ${\approx}2.05\times$ MUDDFormer and $1.27\times$ HyperConnections decode throughput at every length. Note that the training efficiency advantage of SATFormer over ResFormer (Table~3) does not persist at inference; the two are effectively tied (within 2\%), as the linear-path advantage that arises during training is not present in the autoregressive decode loop.
\subsection{General Language Modeling and Zero-Shot Evaluation}

\begin{wraptable}{r}{0.65\textwidth}
    \vspace{-15pt}
    \centering
    \small
    \scshape
    \caption{\textbf{SATFormer lowers ResFormer validation loss at all four evaluated scales.}\protect\footnotemark}
    \label{table4}
    \begin{tabular}{lcccccc}
        \toprule
        \textbf{Variant} & \textbf{TF} & \textbf{DF} & \textbf{MF} & \textbf{HC} & \textbf{RF} & \textbf{SAT} \\
        \midrule
        Small  & 2.707 & 2.704 & 2.658 & 2.670 & 2.674 & \textbf{2.648} \\
        Medium & 2.461 & 2.452 & 2.415 & \textbf{2.410} & 2.429 & 2.415 \\
        Large  & 2.256 & 2.256 & \textbf{2.233} & 2.234 & 2.246 & 2.242 \\
        XL     & 2.152 & N/A    & N/A    & N/A             & 2.150 & \textbf{2.139} \\
        \bottomrule
    \end{tabular}
    \vspace{-5pt}
\end{wraptable}
\footnotetext{TF: Transformer, DF: DenseFormer, MF: MUDDFormer, HC: HyperConnections, RF: ResFormer, SAT: SATFormer.}

We next evaluate whether SATFormer's retrieval and efficiency gains come at the cost of general language-modeling quality. Table~\ref{table4} reports final validation loss, while Table~\ref{table5} reports zero-shot performance across language modeling and downstream benchmarks. These broader evaluations serve as a general-purpose validation of SATFormer beyond the targeted retrieval and efficiency settings.
\begin{table*}[ht]
\centering
\caption{\textbf{SATFormer's clearest language-modeling gains over ResFormer occur at 130M and 1.3B, while downstream differences are scale- and task-dependent.} Best results are \textbf{bolded}. Wiki: WikiText-103; LMB: LAMBADA; Hella.: HellaSwag; Wino.: Winogrande; ARC-e/c: ARC Easy/Challenge; SIQA: Social IQA.}
\label{table5}
\begin{small}
\begin{sc}
\setlength{\tabcolsep}{4.2pt}
\resizebox{\textwidth}{!}{
\begin{tabular}{l | cc | cccccccc | c}
\toprule
\textbf{Model} & \textbf{Wiki.} & \textbf{LMB.} & \textbf{LMB.} & \textbf{PIQA} & \textbf{Hella.} & \textbf{Wino.} & \textbf{ARC-e} & \textbf{ARC-c} & \textbf{SIQA} & \textbf{BoolQ} & \textbf{Avg.} \\
& ppl $\downarrow$ & ppl $\downarrow$ & acc $\uparrow$ & acc $\uparrow$ & acc\_n $\uparrow$ & acc $\uparrow$ & acc $\uparrow$ & acc\_n $\uparrow$ & acc $\uparrow$ & acc $\uparrow$ & $\uparrow$ \\
\midrule
\midrule
\multicolumn{12}{c}{Small: 130M params / 5B tokens} \\
\midrule
Transformer    & 41.81 & 121.31 & 22.26 & 61.21 & 30.57 & 51.85 & 46.68 & 25.51 & 37.00 & \textbf{62.08} & 42.15 \\
DenseFormer      & 41.80 & 134.53 & 21.62 & 61.32 & 30.90 & 50.43 & 49.12 & 23.89 & 35.93 & 60.37 & 41.70 \\
MUDDFormer       & 39.45 & 115.12 & 21.70 & 60.88 & 31.48 & \textbf{52.49} & 48.70 & 23.81 & 36.28 & 57.92 & 41.66 \\
HyperConnections       & 38.45 & 95.54 & 23.71 & 61.31 & 31.63 & 50.98 & 48.44 & \textbf{25.68} & \textbf{37.25} & 55.41 & 41.80 \\
ResFormer        & 39.68 & 106.80 & 23.38 & 60.07 & 31.00 & 51.30 & 48.53 & 23.89 & 35.93 & 61.41 & 41.94 \\
SATFormer       & \textbf{37.98} & \textbf{84.23} & \textbf{25.79} & \textbf{61.97} & \textbf{32.12} & 50.91 & \textbf{49.54} & 25.00 & 36.49 & 59.88 & \textbf{42.71} \\
\midrule
\multicolumn{12}{c}{Medium: 340M params / 10B tokens} \\
\midrule
Transformer    & 29.46 & 59.29 & 27.38 & 65.07 & 36.71 & 50.83 & 52.95 & 24.57 & 38.02 & 62.14 & 44.71 \\
DenseFormer      & 28.92 & 49.68 & 29.17 & 65.40 & 37.01 & 48.15 & 54.42 & 25.68 & 37.87 & 61.01 & 44.84 \\
MUDDFormer       & 27.97 & 49.44 & 27.77 & 65.72 & 37.04 & 51.07 & 54.17 & 26.11 & 37.15 & 59.33 & 44.80 \\
HyperConnections       & \textbf{26.83} & \textbf{35.21} & \textbf{33.86} & 64.85 & \textbf{38.72} & 51.06 & 54.54 & 26.70 & 36.74 & \textbf{61.89} & \textbf{46.04} \\
ResFormer        & 27.84 & 42.38 & 30.47 & 64.74 & 38.03 & \textbf{51.46} & 55.85 & 25.68 & 38.18 & 60.55 & 45.62 \\
SATFormer       & 26.98 & 45.84 & 28.90 & 65.89 & 38.30 & 51.07 & 56.27 & 27.22 & 37.36 & 61.44 & 45.81 \\
\midrule
\multicolumn{12}{c}{Large: 760M params / 20B tokens} \\
\midrule
Transformer    & 21.95 & 24.27 & 36.72 & 67.14 & 44.20 & 50.75 & 61.11 & 30.89 & 37.92 & 60.61 & 48.67 \\
DenseFormer      & 21.77 & 24.57 & 37.16 & 68.17 & 44.28 & 52.17 & 61.70 & 29.52 & 39.15 & 55.35 & 48.44 \\
MUDDFormer       & \textbf{21.07} & 22.30 & 38.11 & 68.28 & 45.53 & 55.64 & 61.74 & \textbf{30.80} & 38.84 & 58.38 & \textbf{49.67} \\
HyperConnections  & 21.14 & \textbf{21.63} & \textbf{38.36} & 67.84 & \textbf{45.93} & \textbf{52.95} & 60.90 & 30.03 & \textbf{39.25} &\textbf{62.17} & \textbf{49.67}\\
ResFormer        & 21.71 & 25.06 & 37.03 & 66.87 & 44.81 & 52.64 & 59.97 & 29.61 & \textbf{39.25} & 60.15 & 48.79 \\
SATFormer       & 21.24 & 24.92 & 36.72 & \textbf{68.34} & 45.25 & 50.67 & \textbf{61.91} & 29.86 & 38.79 & 58.96 & 48.81 \\
\midrule
\multicolumn{12}{c}{XL: 1.3B params / 30B tokens} \\
\midrule
Transformer    & 18.83 & 18.66 & 40.38 & 69.85 & 50.40 & \textbf{54.85} & 65.61 & 32.93 & 39.71 & 56.17 & 51.23 \\
ResFormer        & 18.62 & 18.28 & 40.55 & \textbf{69.36} & 50.02 & 52.88 & 65.48 & \textbf{33.87} & 39.66 & 58.53 & 51.29 \\
SATFormer       & \textbf{18.47} & \textbf{16.44} & \textbf{43.56} & 68.49 & \textbf{51.01} & 54.30 & \textbf{65.69} & 32.59 & \textbf{40.63} & \textbf{59.48} & \textbf{51.96}  \\
\bottomrule
\end{tabular}
}
\end{sc}
\end{small}
\end{table*}

We first focus on the comparison between SATFormer and ResFormer, since this isolates the effect of replacing static value-residual access with input-dependent selective access. SATFormer improves final validation loss over ResFormer at every evaluated scale: 2.648 vs.\ 2.674 at 130M, 2.415 vs.\ 2.429 at 340M, 2.242 vs.\ 2.246 at 760M, and 2.139 vs.\ 2.150 at 1.3B. This suggests that selective access to $V_1$ improves the training objective relative to static value mixing without
requiring dense connectivity. This is the central efficiency advantage of SATFormer: it increases the granularity of cross-layer reuse without substantially increasing the amount of cross-layer connectivity.

One result warrants direct acknowledgment: while SATFormer improves over ResFormer at every evaluated scale, denser alternatives still achieve higher endpoint zero-shot accuracy in some settings. At the Medium scale, HyperConnections outperforms SATFormer in average zero-shot accuracy (46.04 vs. 45.81), and at the Large scale, MUDDFormer and HyperConnections achieve higher average scores (49.67 vs. 48.81). However, SATFormer achieves roughly $1.75$--$1.82\times$ higher throughput than these denser alternatives in our efficiency benchmark. This is consistent with SATFormer’s design intent: it is not meant to uniformly dominate all dense connectivity mechanisms on endpoint accuracy, but to provide a favorable Pareto trade-off between adaptivity, model quality, and training efficiency.

\paragraph{Scaling behavior.}
The scaling results on core language modeling suggest that SATFormer's benefits may depend not only on parameter count, but also on the number of opportunities the model has to reuse $V_1$ across depth. Notably, the XL setting (1.3B parameters) shows SATFormer's strongest zero-shot gains over ResFormer ($+0.67$ average accuracy), consistent with deeper models providing more layers at which selective $V_1$ access can be applied. This interpretation is further supported by the gate analysis in Section~\ref{sec:mechanistic}, where $V_1$ utilization remains sparse in early layers and increases substantially in deeper layers, and by the additional 760M thin-and-long study in Appendix \ref{appendix:b3}, which shows that reallocating capacity toward depth improves SATFormer more than ResFormer. These results suggest that the comparatively muted 760M endpoint result is at least partly configuration-dependent, and that selective access benefits from depth as well as scale. A more complete scaling study separating the effects of depth, width, and training tokens remains an important direction for future work.

\section{Mechanistic Analysis}
\label{sec:mechanistic}

The goal of this section is to test a narrower mechanistic prediction of the architecture: if selective access is doing useful work, the learned gate should not behave like a dense residual shortcut. Instead, access to $V_1$ should be sparse, concentrated in particular heads or layers, and functionally important when perturbed. We focus on three questions: First, is access to $V_1$ sparse or broadly active? Second, which layers are most functionally sensitive to perturbing the learned gate, and how does this compare to where gate activation is largest? Third, do different token categories receive different amounts of access?

\subsection{Sparse and Depth-Dependent Access}
\begin{wrapfigure}{r}{0.55\textwidth}
    \vspace{-5pt}
    \includegraphics[width=0.95\linewidth]{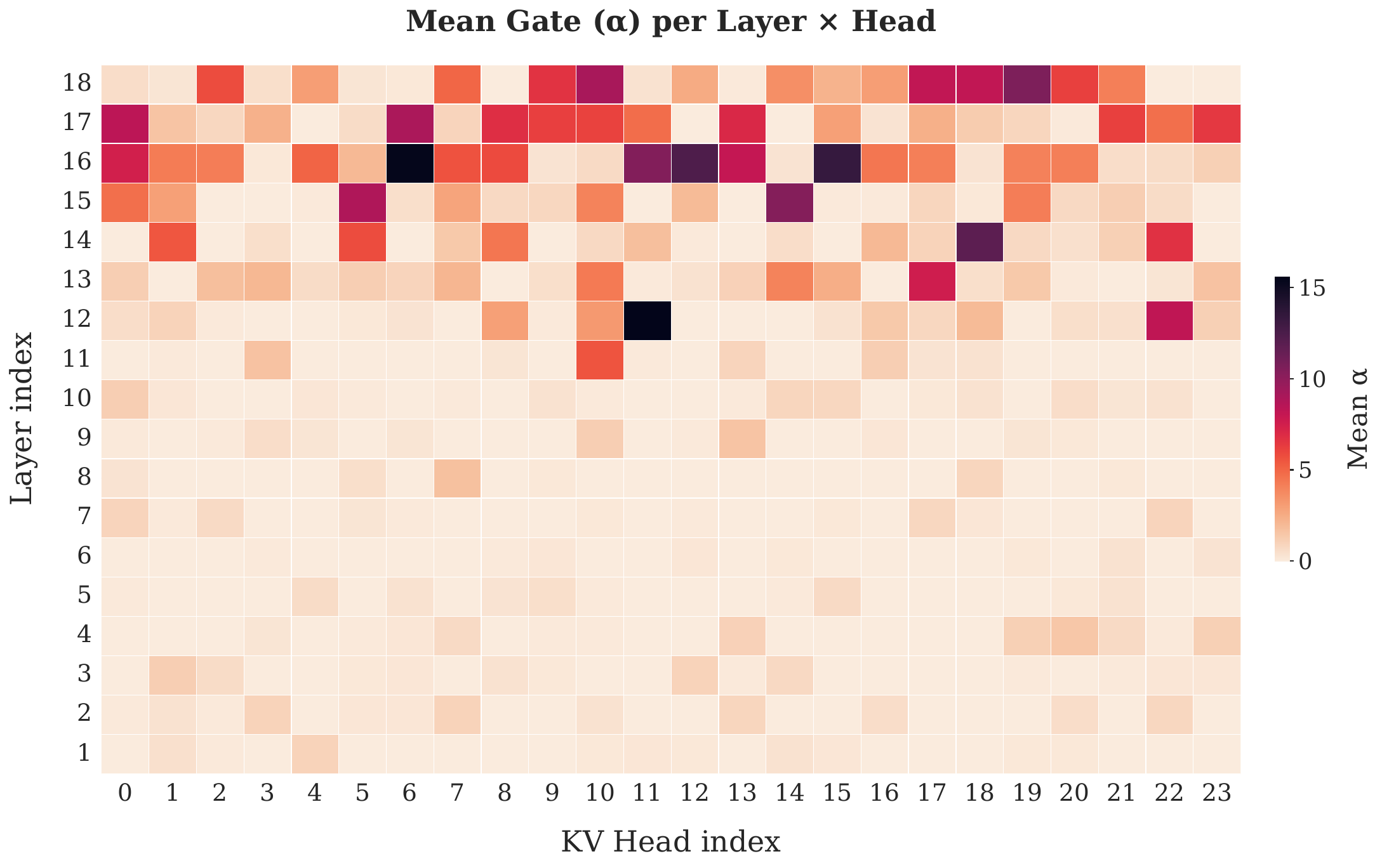}
    \caption{\textbf{SATFormer allocates $V_1$ access to a sparse subset of later-layer heads.}} Mean gate activation across layers and KV heads. Early layers show less access to $V_1$, while some later-layer heads activate sharply, indicating selective rather than uniform early-value reuse.
    \vspace{-35pt}
    \label{figure3}
\end{wrapfigure}

Figure~\ref{figure3} shows the mean gate activation across layers and KV heads. The learned access pattern is highly non-uniform. In the early and middle layers, most heads assign little or no weight to $V_1$, indicating that SATFormer does not simply propagate the first-layer value stream through the network as a dense residual shortcut. Instead, access remains sparse for much of the model depth.
This pattern rules out the simplest interpretation of SATFormer as merely learning a rescaled value residual. If the gate were only compensating for the absence of ResFormer's layer-wise coefficient, access would be distributed broadly across heads within each layer. Instead, the model allocates $V_1$ access sparsely: most heads remain close to inactive, while a smaller set of later-layer heads carries most of the early-value reuse. Thus, SATFormer learns a division of labor across heads and depth, where the early-value pathway is reserved for specialized computations rather than used as a uniform shortcut.

\subsection{Gate Intervention Analysis}

\begin{figure}[htb]
    \centering
    \begin{minipage}{0.48\textwidth}
        \centering
        \includegraphics[width=\linewidth]{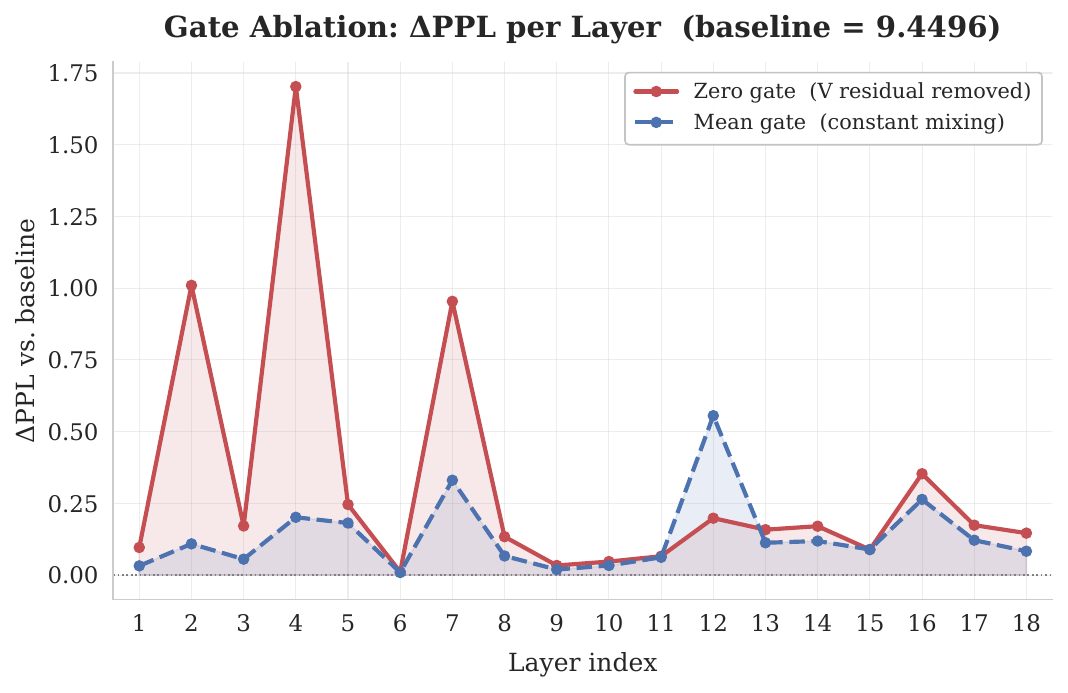}
        \caption{\textbf{Both the $V_1$ pathway and the input-dependent gate structure are functionally load-bearing.} Zeroing or mean-replacing the gate both increase perplexity, with some earlier layers producing disproportionately large effects.}
        \label{figure4}
    \end{minipage}
    \hfill 
    \begin{minipage}{0.48\textwidth}
        \centering
        \includegraphics[width=\linewidth]{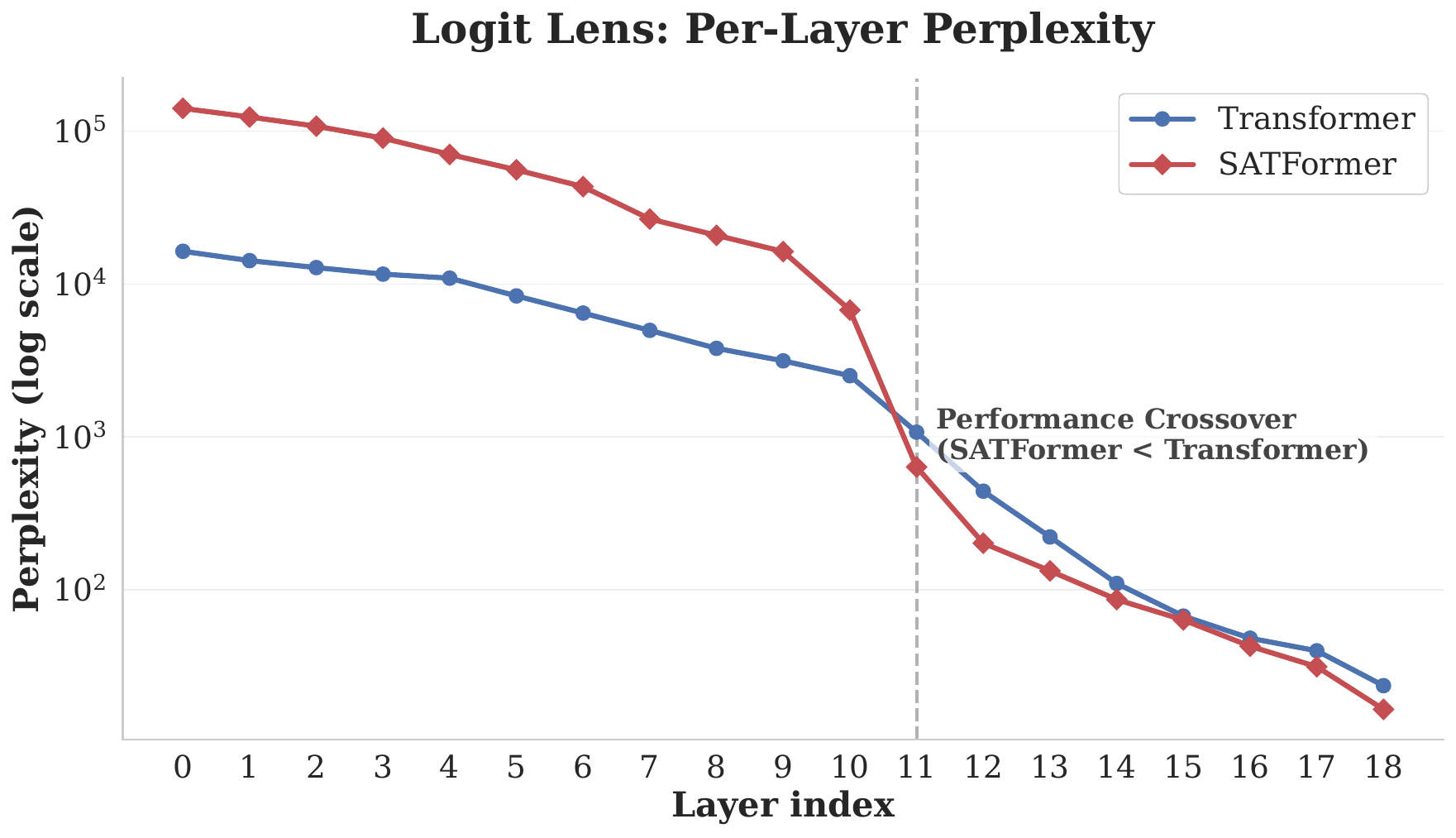}
        \caption{\textbf{SATFormer's logit-lens advantage over the Transformer appears in the same depth range where gate activation increases.} Layer-wise logit-lens perplexity shows a crossover in the later layers where $V_1$ access strengthens; the co-occurrence is suggestive but not by itself causal.}
        \label{figure5}
    \end{minipage}
\end{figure}

As shown in Figure \ref{figure4}, both interventions degrade performance relative to the unablated SATFormer. Zeroing the gate tests whether access to $V_1$ matters at all, while replacing the gate with its mean preserves the average scale of access but removes token- and head-dependent selectivity. The fact that both interventions increase perplexity shows that SATFormer benefits from both components: the early-value pathway itself and the learned structure controlling when that pathway is used.
The intervention results also show that gate magnitude alone is not a complete measure of functional importance. Later layers have the largest average gate activations, but some earlier layers produce large perplexity increases when perturbed. This indicates that small early uses of $V_1$ can affect representations that are amplified or reused later in the network. The logit-lens probe in Figure \ref{figure5} is consistent with this picture: SATFormer initially lags the baseline Transformer, then crosses over in the same later-depth regime where gate activation becomes stronger. Together, the ablation and logit-lens results show that selective access is not only visible in the gate statistics, but also consequential for model predictions. 

\subsection{Category-Specific Access}
\begin{figure}[htb]
    \centering
    \includegraphics[width=0.85\linewidth]{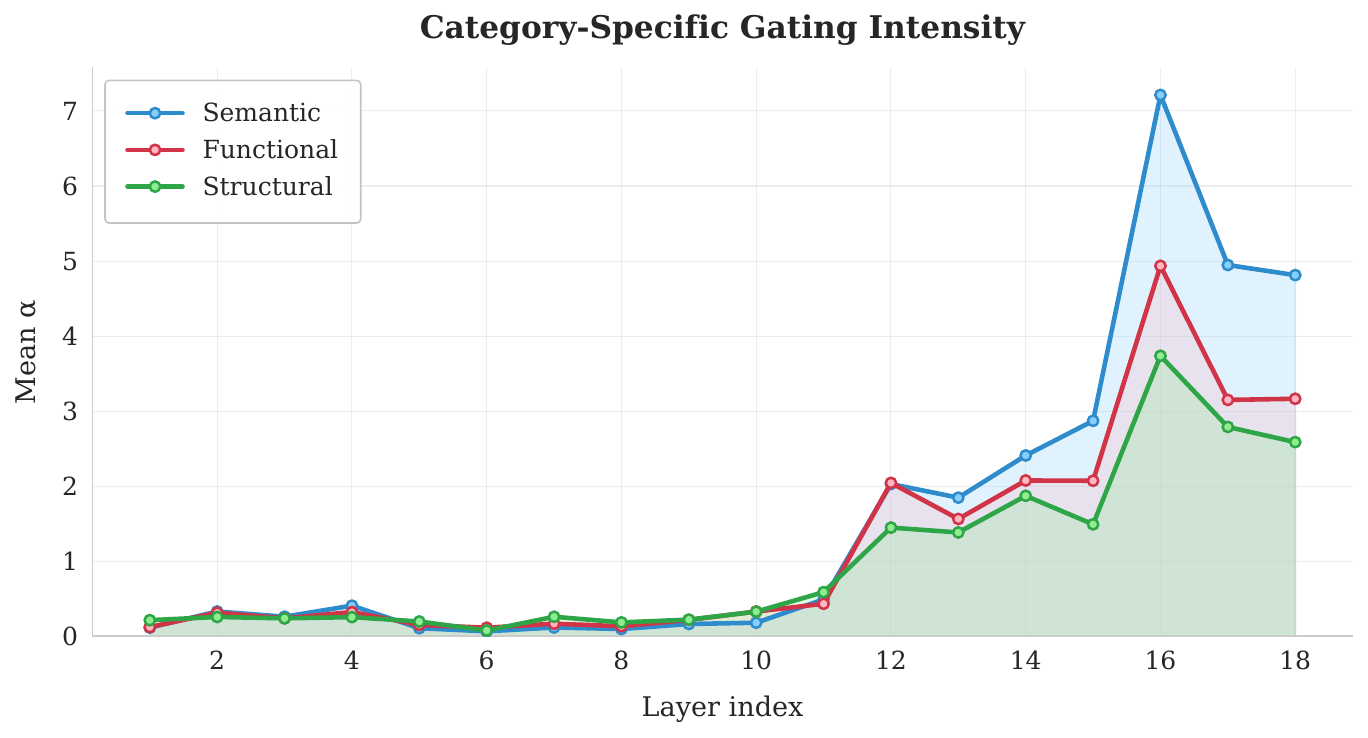}
    \caption{\textbf{Semantic features receive disproportionately stronger $V_1$ access in later layers.} Gating is nearly uniform in the first half of the model, where access is weak overall. In later layers the gap between semantic, functional, and structural categories widens, consistent with selective reuse of content-bearing early representations.}
    \label{figure6}
\end{figure}
Finally, we ask whether the gate treats different kinds of early features differently. Because $V_1$ itself is high-dimensional and not directly labeled, we use sparse autoencoders as a coarse labeling tool: SAE features with similar top-activating tokens are grouped into broad semantic, functional, and structural categories. This analysis is \emph{descriptive}: it characterises which feature categories receive stronger access, but does not causally decompose the source of SATFormer's downstream gains. The causal evidence for the gate's functional role is provided by the intervention experiments in Section~5.2. Details of the SAE training and labeling pipeline are given in Appendix \ref{appendix:c}.

Figure \ref{figure6} shows that category differences are small in early layers, where access to $V_1$ is weak overall. In later layers, however, features labeled as semantic receive higher average gate activation than structural or functional features. This provides a more interpretable view of the sparse late-layer access pattern from Figure \ref{figure3}: the heads that re-access $V_1$ are not only active late in depth, but also tend to assign more access to content-bearing early features. This does not establish that semantic gating causes the retrieval gains in Table 2, but it supports the intended picture of SATFormer as selectively reusing early token information rather than uniformly copying the first-layer value stream.

Across these analyses, SATFormer's learned gate behaves as a selective and functionally consequential access mechanism. Access to $V_1$ is sparse in early layers and concentrated in a subset of later-layer heads; per-layer ablations show that disrupting the gate increases perplexity, with especially large effects in some earlier layers; logit-lens analysis shows that SATFormer begins to outperform the baseline Transformer near the later-layer gating transition; and category-level analysis shows stronger access for semantic than structural features. These findings support the interpretation of SATFormer as a computationally cheap mechanism for targeted early-representation reuse; however, we do not claim full causality and leave fine-grained decomposition as future work.

\section{Limitations}
\label{sec:limitations}

\paragraph{Single-seed training.} All results are from single training runs, so small differences should be interpreted cautiously, particularly the 760M SATFormer--ResFormer validation-loss gap. Multi-seed evaluation remains important future work.
\paragraph{Validation vs. Reasoning.} Related to the above, the gap in training loss between SATFormer and ResFormer is visually small in Figure~\ref{figure2}, while downstream evaluations show larger differences. Small differences in language-modeling can translate unevenly into downstream accuracy, and zero-shot benchmarks at this scale carry substantial evaluation noise of their own. We therefore avoid attributing the downstream gaps to the loss difference, and we rest this part of the argument on the evaluated retrieval benchmarks and the Needle-in-a-Haystack stress test.

\paragraph{Scale and evaluation scope.} Our largest model is 1.3B parameters trained on 30B tokens, and whether the observed trends persist into the 7B+ regime remains unverified. Long-context results rely on YaRN extension applied identically to all architectures rather than on native long-context training, and the benchmark suites available at this scale are narrower than those used for frontier models, as discussed in Section~\ref{sec:niah}. Finally, the SAE-based category analysis is descriptive rather than causal, and inherits the tokenizer and labeling-model constraints detailed in Appendix~\ref{appendix:c}. We additionally do not evaluate multi-step retrieval, tool use, or agentic systems, so the present results should not be interpreted as evidence of improvements in those broader capabilities.

\paragraph{Source-layer and cross-family scope.} We fix the reusable source to $V^{(1)}$ to isolate the access-policy comparison with ResFormer. The current experiments therefore show that selective control improves reuse of the first-layer value in these Llama-style models.

\paragraph{SAE analysis.} The SAE-based categories should be interpreted as a coarse analysis tool rather than a definitive semantic labeling of early representations. The Llama-2 tokenizer produces many subword fragments and byte-level artifacts, which limits the granularity of linguistic categories.  In addition, the labeling pipeline relies on an LLM consistency check and therefore may inherit model-driven biases. For this reason, we use the SAE labels only to support the qualitative conclusion that SATFormer gate activation varies across broad feature categories.
\section{Conclusion}
\label{6 Conclusion}

We introduced SATFormer, a computationally cheap interpretation of value residual learning that replaces static layer-wise access to the first-layer value stream with token- and head-dependent gating. Across 130M to 1.3B parameter models, SATFormer improves validation loss over ResFormer at every evaluated scale and improves average zero-shot accuracy over ResFormer at all four scales, though the 760M zero-shot margin ($0.02$ points) is within single-seed noise; the clearest zero-shot gains are at 130M and 1.3B, including a $+0.67$ point gain at 1.3B. On retrieval-intensive benchmarks, SATFormer achieves the highest average score among all evaluated architectures (29.86 average across TriviaQA, SWDE, SQuAD, NQ, FDA, and DROP), outperforming ResFormer by roughly 1.5 average points and matching MUDDFormer to within $0.01$ points at a fraction of the throughput cost. At the same time, SATFormer maintains throughput and memory usage close to the baseline Transformer and ResFormer, achieving roughly $1.75\times$ higher throughput than HyperConnections and using 31\% less memory. This efficiency profile carries over to inference, where SATFormer's peak memory is indistinguishable from the baseline Transformer at every sequence length tested while sustaining ${\approx}1.87\times$ MUDDFormer prefill throughput at 32K. In the Needle-in-a-Haystack stress test, SATFormer is the only evaluated architecture whose 32K accuracy exceeds its 4K accuracy, with a $+12.7$ point endpoint change. This trajectory is non-monotonic, however, and MUDDFormer achieves higher absolute accuracy at every tested context length. We therefore interpret this result as specific to this single-key retrieval setup rather than as evidence of a general long-context advantage.

These results support a targeted conclusion: early representations can be selectively reused without introducing dense cross-layer connectivity. In our evaluated settings, SATFormer provides a useful design point between static value residuals and the denser cross-layer baselines: it introduces token- and head-dependent control while retaining a computational profile close to ResFormer, and is competitive on the retrieval benchmarks considered here. These findings motivate selective access as a simple and computationally efficient direction for future Transformer architectures, while leaving its behavior in broader long-context and more complex retrieval settings for future study.

\section*{Broader Impacts and Safeguards}

This work studies architectural mechanisms for improving information flow in Transformer language models. Our experiments are conducted at relatively small research scales, and the trained models are intended for controlled empirical analysis rather than deployment. As such, the models evaluated in this work are unlikely to directly pose substantial real-world risks on their own.

At the same time, SATFormer is a general-purpose architectural modification and could be applied to larger language models. Improvements in language-modeling efficiency or performance may therefore contribute to both beneficial and harmful downstream capabilities, depending on how such models are trained, released, and used. Potential risks include misuse of stronger language models for deception, spam, harmful content generation, or other malicious applications. We therefore emphasize that this work should be used in accordance with standard safeguards for language-model development, including careful dataset curation, capability and misuse evaluations, controlled release practices when appropriate, and monitoring for downstream harms.

The intended benefit of this research is to improve the efficiency and interpretability of Transformer architectures. By studying selective access to early representations, we aim to better understand how models reuse information across depth and to identify architectural designs that improve performance without requiring substantially denser or more computationally expensive connectivity.

\bibliographystyle{abbrvnat} 
\bibliography{citations}

\appendix

\section{SATFormer PyTorch Implementation}
\label{appendix:pytorch}
\definecolor{codegreen}{rgb}{0,0.6,0}
\definecolor{codegray}{rgb}{0.5,0.5,0.5}
\definecolor{codepurple}{rgb}{0.58,0,0.82}
\definecolor{backcolour}{rgb}{0.95,0.95,0.92}

\lstdefinestyle{mystyle}{
    backgroundcolor=\color{backcolour},   
    commentstyle=\color{codegreen},
    keywordstyle=\color{magenta},
    numberstyle=\tiny\color{codegray},
    stringstyle=\color{codepurple},
    basicstyle=\ttfamily\footnotesize,
    breakatwhitespace=false,         
    breaklines=true,                 
    captionpos=b,                    
    keepspaces=true,                 
    numbers=left,                    
    numbersep=5pt,                  
    showspaces=false,                
    showstringspaces=false,
    showtabs=false,                  
    tabsize=2
}
\lstset{style=mystyle}

\begin{lstlisting}[language=Python]
import torch
import torch.nn as nn
import torch.nn.functional as F

class SATFormerResidual(nn.Module):
    def __init__(self, hidden_size, num_kv_heads, gate='relu'):
        super().__init__()
        self.num_kv_heads = num_kv_heads
        self.gate = gate
        # Lightweight linear projection: hidden_size -> num_kv_heads
        self.alpha_proj = nn.Linear(hidden_size, num_kv_heads, bias=False)

    def _apply_gate(self, logits):
        if self.gate == 'relu':
            return F.relu(logits)
        elif self.gate == 'sigmoid':
            return torch.sigmoid(logits)
        elif self.gate == 'softmax':
            # Competition across heads; scaled to match unit initialization
            return F.softmax(logits, dim=-1) * self.num_kv_heads
        elif self.gate == 'softmax_sigmoid':
            # Combined across-head competition and per-head gating
            return F.softmax(logits, dim=-1) * torch.sigmoid(logits) * self.num_kv_heads
        elif self.gate == 'tanh':
            return torch.tanh(logits)
        return logits

    def forward(self, v_n, v_1, hidden_states):
        # Compute token-dependent mixing coefficients alpha
        # logits shape: [Batch, Seq, Heads]
        logits = self.alpha_proj(hidden_states.to(v_n.dtype))
        alpha = self._apply_gate(logits).unsqueeze(-1) # [B, T, H, 1]
        
        # Apply mixing: V'_n = V_n + alpha(h) * V_1
        return v_n + alpha * v_1.to(v_n.dtype)

# --- Integration into Model Forward Pass ---
# In SATFormerModel.forward():
V1 = None
for layer_idx, layer in enumerate(self.layers):
    # Each layer returns (hidden_states, ..., v_base)
    # v_base is the raw V projection from the current layer's Attention module
    outputs = layer(hidden_states, V1=V1)
    hidden_states = outputs[0]
    v_base = outputs[-1]

    # Step 1: Capture V1 from Layer 0
    if layer_idx == 0:
        V1 = v_base  # Store raw projection for all subsequent layers

    # Step 2: Propagation
    # For layers n > 0, the 'V1' variable now contains the stored projection.
    # It is passed into the Attention module of the next layer via 'layer(...)'.
\end{lstlisting}

\section{Ablation Study}
\subsection{Gating Analysis}
\label{appendix:b1}
\begin{figure}[htb]
    \centering
    \includegraphics[width=0.9\linewidth]{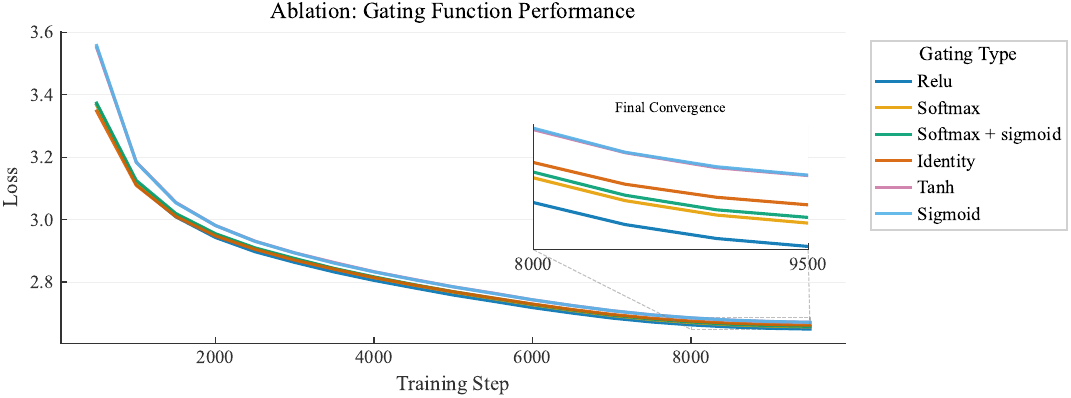}
    \caption{\textbf{ReLU provides the lowest and most stable loss among the tested gates.} Gating-function ablation at 130M parameters and 5B training tokens.}
    \label{figure7}
\end{figure}
To identify the optimal mechanism for token-dependent value mixing, we evaluate several candidate gating functions for the coefficient $\alpha_{t,j}^{(n)}$.We tested six specific gating behaviors (Note: we use the standard PyTorch Kaiming uniform initialization for all linear layers):
\begin{enumerate}
    \item \textbf{ReLU (Default)}: Produces sparse, non-negative weights, allowing the model to completely zero out the connection if/when $V_1$ is redundant.
    \item \textbf{Sigmoid}: Provides a smooth, bounded $[0, 1]$ probability-like weight for each head.
    \item \textbf{Softmax}: Computes a convex combination across the head dimension, forcing heads to ''compete'' for a shared $V_1$ budget (scaled by $N_{kv}$ to maintain signal parity).
    \item \textbf{Softmax + Sigmoid}: A hybrid approach combining across-head competition with an individual per-head on/off gate.
    \item \textbf{Tanh}: Allows for subtractive mixing ($[-1, 1]$), enabling the model to suppress features in the current layer using the original input.
    \item \textbf{Identity}: A linear baseline ($Wh$) used to verify the necessity of non-linear activation.
\end{enumerate}

As shown in Figure \ref{figure7}, the ReLU and Softmax gates demonstrate the most stable convergence and lowest final loss (although ReLU maintains a significant lead over Softmax). Surprisingly, Sigmoid and Tanh perform the worst, possibly suggesting that bounded gating limits the expressivity enabled with value mixing. Overall, ReLU is simultaneously the simplest and strongest performing gating function, exhibiting a favorable inductive bias for enabling deeper expressivity with value mixing.

\subsection{Selective Value Access Improves Gated-Attention Perplexity}
\label{appendix:bgate}

Gated Attention \citep{qiu2026gated} applies a head-specific sigmoid gate to the attention output before the output projection. SATFormer instead gates the reuse of $V^{(1)}$ inside the value stream. The two mechanisms therefore control different quantities: output gating determines how strongly the current attention computation enters the residual stream, whereas selective value access determines which early representation that computation can reuse. To test whether output gating already makes value-residual gating redundant, we compare Gated Attention against the same architecture augmented with SATFormer at the 130M scale, using matched data, optimization, and hardware settings.

\begin{table*}[ht]
  \centering
  \caption{\textbf{Adding SATFormer to Gated Attention lowers language-modeling perplexity, substantially on LAMBADA and modestly on WikiText-103.} WikiText-103 perplexity decreases by 0.77 (2.0\%) and LAMBADA perplexity by 21.82 (21.5\%); downstream task effects are mixed, leaving the average accuracy nearly unchanged. Best results are \textbf{bolded}.}
  \label{tab:gated_attention_results}
  \begin{small}
  \setlength{\tabcolsep}{4.2pt}
  \resizebox{\textwidth}{!}{
  \begin{tabular}{l | cc | cccccccc | c}
  \toprule
  \textbf{Model}
  & \textbf{Wiki.}
  & \textbf{LMB.}
  & \textbf{LMB.}
  & \textbf{PIQA}
  & \textbf{Hella.}
  & \textbf{Wino.}
  & \textbf{ARC-e}
  & \textbf{ARC-c}
  & \textbf{SIQA}
  & \textbf{BoolQ}
  & \textbf{Avg.} \\
  & ppl $\downarrow$
  & ppl $\downarrow$
  & acc $\uparrow$
  & acc $\uparrow$
  & acc\_n $\uparrow$
  & acc $\uparrow$
  & acc $\uparrow$
  & acc\_n $\uparrow$
  & acc $\uparrow$
  & acc $\uparrow$
  & $\uparrow$ \\
  \midrule
  Gated Attention
  & 38.09
  & 101.60
  & 23.25
  & \textbf{62.62}
  & \textbf{31.91}
  & 51.54
  & \textbf{50.97}
  & \textbf{24.83}
  & \textbf{37.82}
  & 60.89
  & \textbf{42.98} \\
  Gated Attention + SATFormer
  & \textbf{37.32}
  & \textbf{79.78}
  & \textbf{26.14}
  & 61.86
  & 31.58
  & \textbf{52.17}
  & 49.41
  & 24.40
  & 36.59
  & \textbf{61.44}
  & 42.95 \\
  \bottomrule
  \end{tabular}
  }
  \end{small}
\end{table*}

The language-modeling effect is consistent in direction across the two perplexity datasets, and large on LAMBADA. Adding selective value access reduces WikiText-103 perplexity from 38.09 to 37.32 (an absolute reduction of 0.77, or 2.0\%) and LAMBADA perplexity from 101.60 to 79.78 (an absolute reduction of 21.82, or 21.5\%). LAMBADA accuracy also rises by 2.89 points, from 23.25 to 26.14. These gains indicate that an output gate does not replace access to an early value source: even after the attention block is gated, controlling which value information enters that block remains useful.

The downstream picture is deliberately more qualified. Gated Attention + SATFormer improves Winogrande and BoolQ but declines on PIQA, HellaSwag, ARC-Easy, ARC-Challenge, and SocialIQA, producing an effectively unchanged average (42.95 versus 42.98). We therefore use this experiment to support a narrow complementarity claim: SATFormer's selective value pathway adds language-modeling value on top of output gating, especially on LAMBADA perplexity, but the present 130M result does not establish a uniform downstream advantage. Larger-scale replication would be needed to determine whether the perplexity gains translate more consistently into task accuracy.

\subsection{Depth vs. Width Study}
\label{appendix:b3}

The main 760M results show that SATFormer improves both validation loss and zero-shot language modeling over ResFormer, however its advantage is less pronounced than at the 130M, 340M, and 1.3B scales. To test whether this behavior reflects the architecture itself or the particular width vs. depth allocation used at 760M, we ran an additional matched-scale experiment using a ``thin-and-long'' configuration (1536 $d_{model}$, 24 $L$, 16 $n_{heads}$, 4352 $d_{ff}$). This configuration reallocates capacity toward depth while remaining in the same approximate parameter regime.
\begin{wraptable}{r}{0.55\textwidth}
\vspace{-5pt}
\centering
\caption{\textbf{Reallocating capacity toward depth improves SATFormer more than ResFormer.} Final validation loss at the matched 760M scale.}
\begin{small}
\begin{sc}
\begin{tabular}{lc}
\toprule
\textbf{Model} & \textbf{Validation Loss} \\
\midrule
ResFormer-Large-Thin & 2.2396 \\
ResFormer-Large      & 2.2461 \\
SATFormer-Large-Thin & \textbf{2.2313} \\
SATFormer-Large      & 2.2424 \\
\bottomrule
\end{tabular}
\end{sc}
\end{small}
\vspace{-5pt}
\label{table6}
\end{wraptable}

While both value-residual architectures benefit from the deeper allocation, SATFormer benefits more strongly. In Table \ref{table6}, ResFormer improves from 2.2461 to 2.2396, whereas SATFormer improves from 2.2424 to 2.2313, corresponding to approximately a 1.7$\times$ larger reduction for SATFormer. The resulting SATFormer-Large-Thin model also outperforms the corresponding ResFormer-Large-Thin model by 0.0083 validation loss. 

\begin{table*}[ht]
\centering
\caption{\textbf{The thin-and-long allocation reverses the original 760M average-accuracy ordering in SATFormer's favor.} Zero-shot language-modeling and downstream performance.}
\begin{small}
\begin{sc}
\setlength{\tabcolsep}{4.2pt}
\resizebox{\textwidth}{!}{
\begin{tabular}{l | cc | cccccccc | c}
\toprule
\textbf{Model} & \textbf{Wiki.} & \textbf{LMB.} & \textbf{LMB.} & \textbf{PIQA} & \textbf{Hella.} & \textbf{Wino.} & \textbf{ARC-e} & \textbf{ARC-c} & \textbf{SIQA} & \textbf{BoolQ} & \textbf{Avg.} \\
& ppl $\downarrow$ & ppl $\downarrow$ & acc $\uparrow$ & acc $\uparrow$ & acc\_n $\uparrow$ & acc $\uparrow$ & acc $\uparrow$ & acc\_n $\uparrow$ & acc $\uparrow$ & acc $\uparrow$ & $\uparrow$ \\
\midrule
ResFormer  & 20.73 & \textbf{21.81} & \textbf{39.01} & 67.85 & \textbf{46.61} & 51.54 & 61.32 & 31.06 & 39.10 & \textbf{59.76} & 49.53 \\
SATFormer  & \textbf{20.42} & 22.04 & 38.19 & \textbf{68.17} & 46.58 & \textbf{52.64} & \textbf{63.30} & \textbf{31.74} & \textbf{39.36} & 58.47 & \textbf{49.81} \\
\bottomrule
\end{tabular}
}
\end{sc}
\end{small}
\label{table7}
\end{table*}

The zero-shot results in Table \ref{table7} show a similar reversal: in the original 760M configuration, ResFormer narrowly outperformed SATFormer in average accuracy, but under the thin-and-long allocation SATFormer takes a larger step forward, improving to 49.81 average accuracy compared with 49.53 for ResFormer. These results support the interpretation that SATFormer benefits disproportionately from depth. This is consistent with the main mechanistic analysis, where gate activation is sparse in early layers and becomes stronger and more head-specialized in later layers. The thin-and-long study suggests that the weaker relative 760M endpoint result is not evidence against the selective-access mechanism, but rather reflects sensitivity to the allocation of parameters between depth and width.

\section{Additional Mechanistic Interpretability}
\label{appendix:c}

This appendix provides additional details for the interpretability analysis in Section~\ref{sec:mechanistic}. The main paper uses three pieces of evidence: sparse depth-dependent gating, logit-lens behavior, and category-specific gate activation. Here, we describe how the token-feature categories were constructed and provide supplementary analyses showing that the broad content of $V_1$ is similar across models, while SATFormer changes how those early features are accessed.

\subsection{SAE Training and Feature Labeling}
\label{appendix:c1}
\begin{table}[htb]
\centering
\caption{\textbf{The $V_1$ probe separates broad content-bearing, functional, and structural feature classes.} Taxonomy used for the category-specific gate analysis.}
\label{table8}
\begin{tabularx}{\textwidth}{@{} l X l @{}}
\toprule
\textbf{Feature Category} & \textbf{Description} & \textbf{Examples} \\ \midrule
Semantic Root Fragments & Core meaning-bearing subwords & \textit{struct, trans, ology} \\
Entity Marker Fragments & Capitalized fragments often used in proper names & \textit{Mc, San, Al} \\
Morphological Suffixes & Subwords that modify parts of speech & \textit{ing, ed, s, ly} \\
Logic Quantifier Keywords & Connectives and quantifiers that direct sentence flow & \textit{if, however, all, but} \\
Functional Leading Space & Tokens that represent the start of a new word & \textit{The, Apple, Run} \\
Numerical Digit Fragments & Numbers, decimals, or ordinal indicators & \textit{12, 4, .5, th} \\
Structural Punctuation & Symbols used for syntax, grouping, or boundaries & \textit{periods, colons, brackets, quotes} \\
Multilingual Unicode & Valid non-ASCII scripts & \textit{$\alpha$, \'e, \~n, \ss} \\
Code Formatting Symbols & Programming syntax or whitespace/indentation & \textit{braces, newline markers, ++} \\
Byte-Level Fallbacks & Raw UTF-8 bytes or non-standard characters & \texttt{<0x0A>} \\
\bottomrule
\end{tabularx}
\end{table}
To obtain coarse categories for $V_1$ features, we train Sparse Autoencoders (SAEs)~\citep{cunningham2023sparse} on the first-layer value activations of the 760M-parameter Transformer and SATFormer models. The SAEs use a Top-$K$ activation function with $k=32$ and an expansion factor of 16. We train both models for 100M tokens using Adam with learning rate $5 \times 10^{-4}$ and a cosine annealing schedule. The resulting SAEs achieve Fraction of Variance Explained above 99\%, indicating that the sparse codes preserve most of the variance in the underlying activation space.

We use the SAEs only to construct a coarse probe of which types of early features receive stronger gate activation. The goal is not to provide a complete semantic decomposition of $V_1$, but to distinguish broad classes of early representations that may be accessed differently by SATFormer.

After training, we identify and categorize high-confidence features using a three-stage automated pipeline:

\begin{enumerate}
    \item \textbf{Density filtering and MonoScore ranking.}
    We remove dead or ultra-sparse features with fewer than 10 activations, as well as overly dense features with activation density above $0.5\%$. For each remaining feature, we compute a MonoScore based on the similarity structure of its top activating tokens.

    \item \textbf{LLM consistency scoring.}
    We select the top 20\% of features by MonoScore for qualitative evaluation. Using Qwen3-30B-A3B-Instruct-2507-FP8~\citep{yang2025qwen3} served with vLLM~\citep{kwon2023efficient}, we perform a consistency check over two non-overlapping sets of highly activating tokens. A feature is retained only if both subsets receive a feature-purity score of at least 7 out of 10.

    \item \textbf{Taxonomic categorization.}
    Retained features are assigned to a predefined taxonomy of linguistic and structural categories. The taxonomy is designed to capture broad distinctions relevant to early-token representations, including semantic fragments, structural tokens, code-related tokens, and tokenizer artifacts.
\end{enumerate}

For visual clarity, we aggregate the fine-grained SAE feature labels in Table~\ref{table8} into three broader categories used in the main-text analysis:
\begin{enumerate}
    \item \textbf{Semantic:} Semantic Root Fragments, Entity Marker Fragments, Logic Quantifier Keywords, Morphological Suffixes, and Functional Leading Space.
    \item \textbf{Functional:} Multilingual Unicode, Byte-Level Fallbacks, and Numerical Digit Fragments.
    \item \textbf{Structural:} Structural Punctuation and Code Formatting Symbols.
\end{enumerate}

\subsection{Supplementary Gate Statistics}
\label{appendix:c2}
\begin{figure}[htb]
    \centering
    \includegraphics[width=0.85\linewidth]{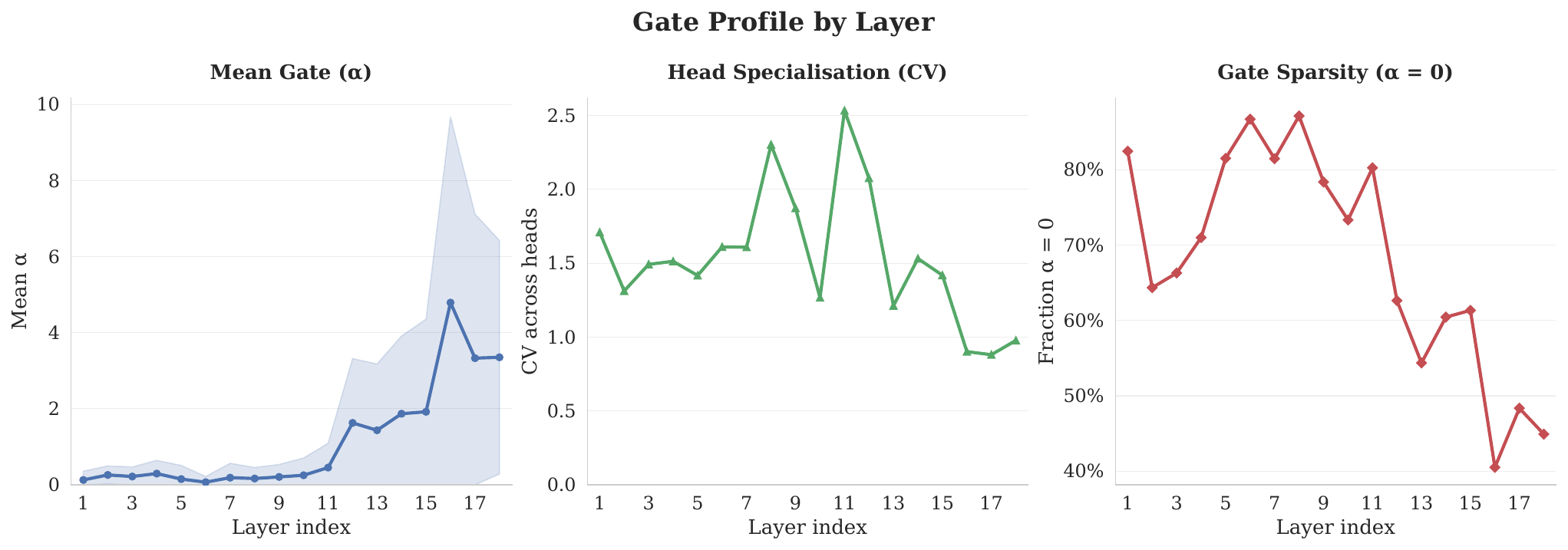}
    \caption{\textbf{Gate access becomes stronger, less sparse, and more head-specialised with depth.} Mean gate activation increases in later layers while gate sparsity decreases and head-level CV grows, providing an aggregate view consistent with the heatmap in Figure~3.}
    \label{figure8}
\end{figure}
\begin{wrapfigure}{r}{0.45\linewidth}
    \vspace{-10pt}
    \centering
    \includegraphics[width=0.95\linewidth]{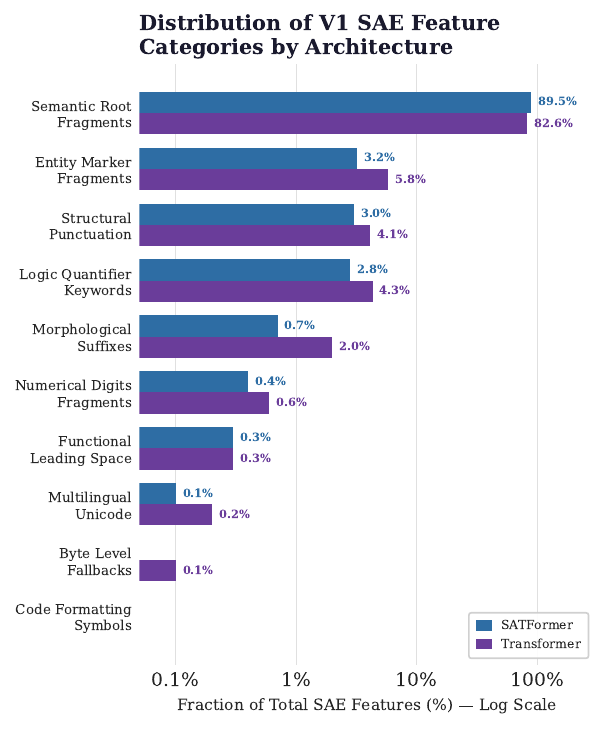}
    \caption{\textbf{SATFormer changes access more than content.} Transformer and SATFormer have similar coarse $V_1$ feature-category distributions.}
    \label{figure9}
    \vspace{-30pt}
\end{wrapfigure}
The heatmap in the main paper shows that large gate activations are concentrated in a subset of later-layer heads. Figure~\ref{figure8} provides a complementary aggregate view of the same behavior. Mean gate activation remains low in early layers and increases in later layers, while the fraction of exactly zero gate values decreases. At the same time, the coefficient of variation across heads increases, indicating that later-layer access is not only stronger but also more head-specialized.

This supports the interpretation in Section~\ref{sec:mechanistic}: SATFormer does not use $V_1$ as a uniform shortcut. Instead, the model learns a sparse access pattern that becomes stronger and more differentiated across heads in deeper layers.

\subsection{Feature Distribution in $V_1$}
\label{appendix:c3}
We also compare the distribution of automatically labeled $V_1$ features across the baseline Transformer and SATFormer. SATFormer produces a larger number of automatically verified monosemantic features under our pipeline, with 3,939 retained features compared to 3,261 for the Transformer. We interpret this as evidence that the accessible feature structure differs between the models, but not as a direct measure of representational richness. This metric can be affected by SAE training dynamics, activation geometry, tokenizer artifacts, and the automated labeling procedure.
Figure~\ref{figure9} shows that the broad category distribution of labeled $V_1$ features is similar across the two architectures. This is important for the main analysis: SATFormer's category-specific gate behavior does not appear to arise from a large shift in the coarse content of $V_1$. Instead, the evidence is more consistent with the view that SATFormer primarily changes how early features are accessed through depth.
\end{document}